\documentclass{article}
\usepackage{spconf}
\usepackage{graphicx}
\usepackage{amsmath}
\usepackage{amssymb}
\usepackage{subfigure}
\usepackage{algorithm}
\usepackage{algorithmicx}
\usepackage{algpseudocode}
\usepackage{setspace}
\usepackage{longtable}
\usepackage{diagbox}
\usepackage{multirow}
\usepackage{color}
\usepackage{url}
\usepackage{stfloats}

\usepackage{color}
\usepackage{amsmath}
\def\eg{\emph{e.g.}}

\def\CD{\cal{D}}

\def\I{\mathbf{I}}

\def\L{\mathbf{L}}

\title{LandmarkGAN: Synthesizing Faces from Landmarks}
\name{Pu Sun$^{1*}$, Yuezun Li$^{2*}$, Honggang Qi$^1$ and Siwei Lyu$^2$ \thanks{* The authors contribute equally.}}
\address{$^1$ University of Chinese Academy of Sciences, China \\
$^2$ University at Buffalo, State University of New York, USA}
\begin{document}

\maketitle
% \twocolumn[{%
% \renewcommand\twocolumn[1][]{#1}%
% \maketitle
% \vspace{-0.3cm}
% \begin{center}
%     \centering
%     \includegraphics[width=0.95\textwidth]{figure/overview.pdf}
%     \vspace{-0.45cm}
%     \captionof{figure}{ \small Visual examples generated by our method. The first row is the input facial landmarks with mouth and head orientation edited (red marks in left and right figure) and the other two rows are synthesized faces of target identity using facial landmarks as input, which showcase the facial expressions and orientations are greatly retained.}
%     \label{fig:teaser}
% \end{center}%
% }]

\begin{abstract}
Face synthesis is an important problem in computer vision with many applications. In this work, we describe a new method, namely LandmarkGAN, to synthesize faces based on facial landmarks as input. Facial landmarks are a natural, intuitive, and effective representation for facial expressions and orientations, which are independent from the target's texture or color and background scene. Our method is able to transform a set of facial landmarks into new faces of different subjects, while retains the same facial expression and orientation. Experimental results on face synthesis and reenactments demonstrate the effectiveness of our method. 

% The implementation of our method can be accessed here: {\tt hidden-for-review}\footnote{Our code will be made public after review. We also provide a GUI interface for editing faces. The demo of the GUI interface can be found in supplementary materials.}.
% \dots
\keywords{Face synthesis, GAN}
\end{abstract}

%%%%%%%%%%%%%%%%%%%%%%%%%%%%%%
%%%%%%%%%%%%%%%%%%%%%%%%%%%%%%
\vspace{-0.3cm}
\section{Introduction}
\vspace{-0.3cm}
Creating realistic images of human faces, as an important problem in computer vision with many practical applications, has recently received a lot of attentions \cite{denton2015deep,arjovsky2017wasserstein,liu2017unsupervised,isola2017image,CycleGAN2017,chen2018cartoongan,choi2018stargan,kim2018deep,wu2018reenactgan,karras2018progressive,karras2019style,zakharov2019few,ha2020marionette,karras2020analyzing}. The general approach of face synthesis is to use a generator, usually in the form of a deep neural network, which takes an input control variable and converts it to a face image. The early face synthesis methods \cite{denton2015deep,arjovsky2017wasserstein,karras2018progressive} are based on Generative Adversarial Networks (GANs) \cite{goodfellow2014generative}, which use random noise as the input control methods. Although highly realistic human face images are generated using these methods, they have a major limitation: the user has little control over the identity and facial attributes such as expression and orientation in the synthesized faces. Face style transfer methods \cite{liu2017unsupervised,isola2017image,CycleGAN2017,chen2018cartoongan,choi2018stargan,karras2019style}  generate new face images by incorporating the style transferred from other domain to the source face image instead of input random noise. Subsequently, face reenactment methods \cite{kim2018deep,wu2018reenactgan,zakharov2019few,ha2020marionette} take face images of a source identity as the system input, and generate faces of a different target identity preserving the facial expression of the input. 

% As such, face reenactment methods predicate on methods that can disentangle identity related information from facial expressions and orientations. 

% \begin{figure*}[t]
%   \centering
%   \includegraphics[width=0.8\linewidth]{figure/teaser.pdf}
%   \vspace{-0.5cm}
%   \caption{ \small Visual examples generated by our method. Our method generates face images of target identity using facial landmarks as input, while greatly retains the facial expressions and orientations. }
%   \label{fig:demo}
% ~\vspace{-2em} \end{figure*}

\begin{figure}
    \centering
    \includegraphics[width=\linewidth]{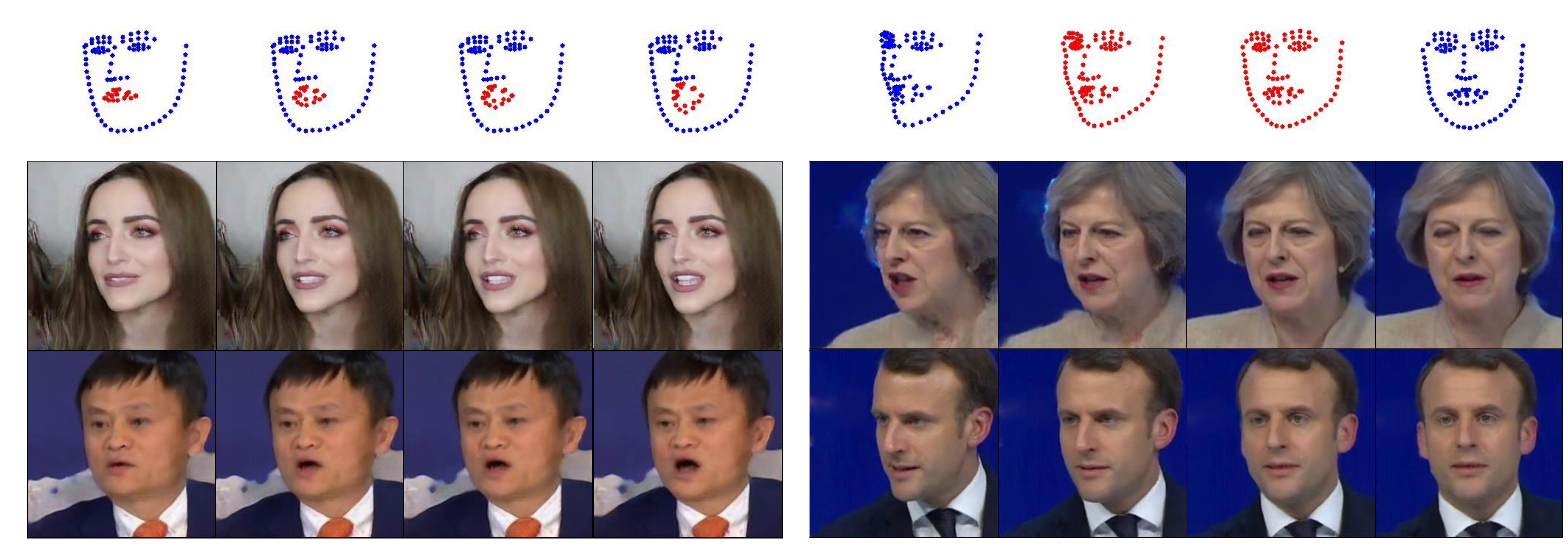}
    \vspace{-0.85cm}
    \caption{\small Visual examples generated by our method. The first row is the input facial landmarks with mouth and head orientation edited (red marks in left and right figure) and the other two rows are synthesized faces of target identity using facial landmarks as input, which showcase the facial expressions and orientations are greatly retained.}
    \label{fig:teaser}
    \vspace{-0.5cm}
\end{figure}

In this work, we describe a new method, known as LandmarkGAN, to synthesize faces only using facial landmarks. Facial landmarks correspond to important locations of facial parts (tips and middle points of eyes, nose, mouth, and eye brows) and contours. The facial landmarks can be reliably detected from input images using state-of-the-art algorithms \cite{sun2019deep,qian2019aggregation}. As the extraction of facial landmarks discard texture and color of the individual faces and any non-face backgrounds, they provide a natural low-dimensional representation for synthesizing faces. 
% Compared to using Action Units (AU) \cite{pumarola2018ganimation,tripathy2020icface}, which are a set of facial muscle movement corresponding to the emotion, facial landmarks are more stable and contain both expression and orientation attributes, which thereby can be used to synthesize faces. 
Moreover, the facial landmarks are structural and human interpretable compared to other signals such as Action Units (AU) \cite{pumarola2018ganimation,tripathy2020icface}, which enables direct editing to generate faces with modified expressions and orientations. Note many previous works achieve the face synthesis or reenactment assisted with the guidance such as facial landmarks, yet few of them focus on synthesizing faces solely based on facial landmarks. The work \cite{di2018gp} synthesizes faces from facial landmarks. However, it only focuses on persevering the genders of generated images instead of identity switching, facial expressions and orientations, which therefore hardly to be applied in reenactment. Fig.\ref{fig:teaser} shows two groups of visual examples generated by our method. We edit the facial landmarks to synthesize corresponding target faces\footnote{We provide a GUI for editing faces. The demo can be found here \url{https://drive.google.com/file/d/1gd_vgCqEULeWt4DMjvXuqqtLTxNQMY6n/view?usp=sharing}.}. The target identity is randomly selected from CelebV dataset \cite{wu2018reenactgan}. 

The proposed face synthesis model has two components. The first is a {\em landmark converter}, which takes an auto-encoder structure to convert the input facial landmarks to those of the target. The converted facial landmarks are then fed to a {\em target-specific landmark-to-face (TL2F) generator}, which is an up-sampling convolutional neural network, to create the face image of target identity incorporating the facial expression and orientation in the facial landmarks. We then describe a new fully differentiable landmark detector to enable landmark transferring consistently. Our model is trained by jointly optimizing the parameters in landmark converter and TL2F generator.

Experimental results show that our method can synthesize face images of target identity with high visual quality and varying facial expressions and orientations. We also implement a face reenactment system based on our method, where the input facial landmarks are extracted from the input faces. When compared with state-of-the-art face reenactment methods, our method achieves competitive performance with improved qualitative and quantitative evaluation results.

\begin{figure}[t]
    \centering
    \includegraphics[width=\linewidth]{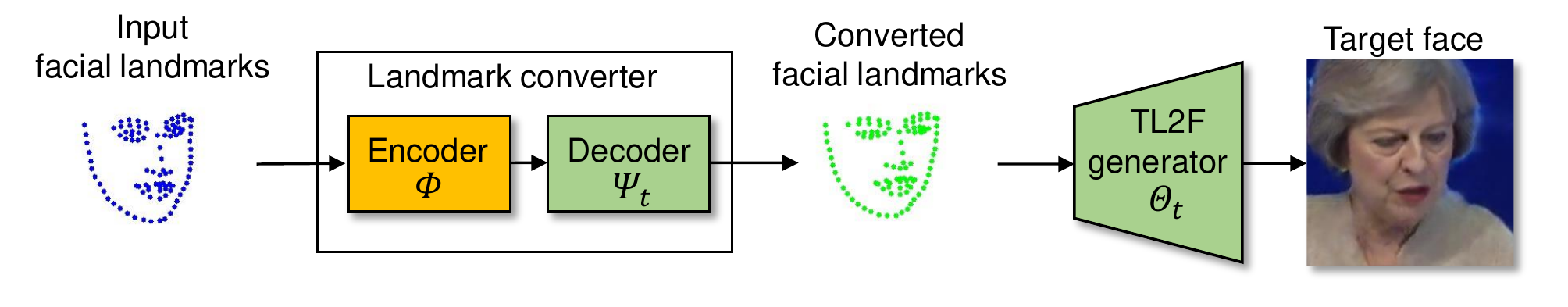}
    \vspace{-1cm}
    \caption{ \small Overview of our method to synthesize a target face.}
    \label{fig:Network_pipeline}
~\vspace{-2em}\end{figure}

% \begin{figure*}[t]
%     \centering
%     \includegraphics[width=0.8\linewidth]{figure/structure.pdf}
%     \vspace{-0.4cm}
%     \caption{ \small Detailed architecture of landmark converter and TL2T generator.}
%     \label{fig:structure}
% ~\vspace{-2em} \end{figure*}

\vspace{-0.3cm}
\section{Methodology}
\vspace{-0.3cm}
\subsection{Model Structure}
\vspace{-0.3cm}
The overall structure of our method is illustrated in Fig.\ref{fig:Network_pipeline}. When the input is the face of a source identity, one can use any off-the-shelf facial landmark detection methods, such as \cite{bulat2017far,sun2019deep} to extract landmarks and use them as inputs for landmark-based face reenactment. In the following, we use ${\L}_{s}$ and ${\I}_{s}$ to denote the input facial landmarks and corresponding image of source identity $s$, respectively. 

The (x,y)-coordinates of the input facial landmarks are first converted to a vector and input to the {\em landmark converter}. The landmark converter has an auto-encoder structure, which is formed by an encoder and a decoder, both are lightweight neural networks consisting of five fully connected (FC) layers. The encoder, subsequently denoted as $\Phi$, is shared across all identities, and converts the input landmark into a latent feature that is identity-neutral but preserve essential facial expressions and orientations. On the other hand, the decoder is {\em target-specific}, which reconstructs landmarks specific to the shape and geometry of its corresponding target from the latent feature. We subsequently denote the decoder in the landmark converter as $\Psi_t$ for target $t$. With a set of facial landmarks for source $s$, ${\L}_{s}$, as input, the converted facial landmarks of target $t$, ${t} \neq s$ is obtained as $\bar{{\L}}_{t} = \Psi_{t}(\Phi({\L}_{s}))$. 

The converted landmarks are then used as the input to a {\em target-specific landmark-to-face (TL2F) generator}, which synthesizes the face image of target identity corresponding to the facial expression and head orientations  represented in the input facial landmarks. The TL2F generator consists of two FC layers and six upscale blocks. The FC layers transform the converted landmarks to a feature vector, which is then reshaped to a feature map. The upscale block is a set of operations that upsamples the input feature map by scale $2$ in width and height. In detail, the upscale block contains a convolutional layer which increases the channel of input feature map, and a PixelShuffle layer \cite{shi2016real} which upsamples the feature map by shifting the elements in channel dimension to width and height dimension. In what follows, we will use  $\Theta_{t}$ to denote the TL2F generator for target $t$. With the converted facial landmarks $\bar{{\L}}_{t}$, the face image of target $t$ synthesized by $\Theta_{t}$ is represented as $\bar{{\I}}_{t} = \Theta_{t}(\bar{\L}_{t})$.

\begin{figure*}[t]
    \centering
    \includegraphics[width=0.8\linewidth]{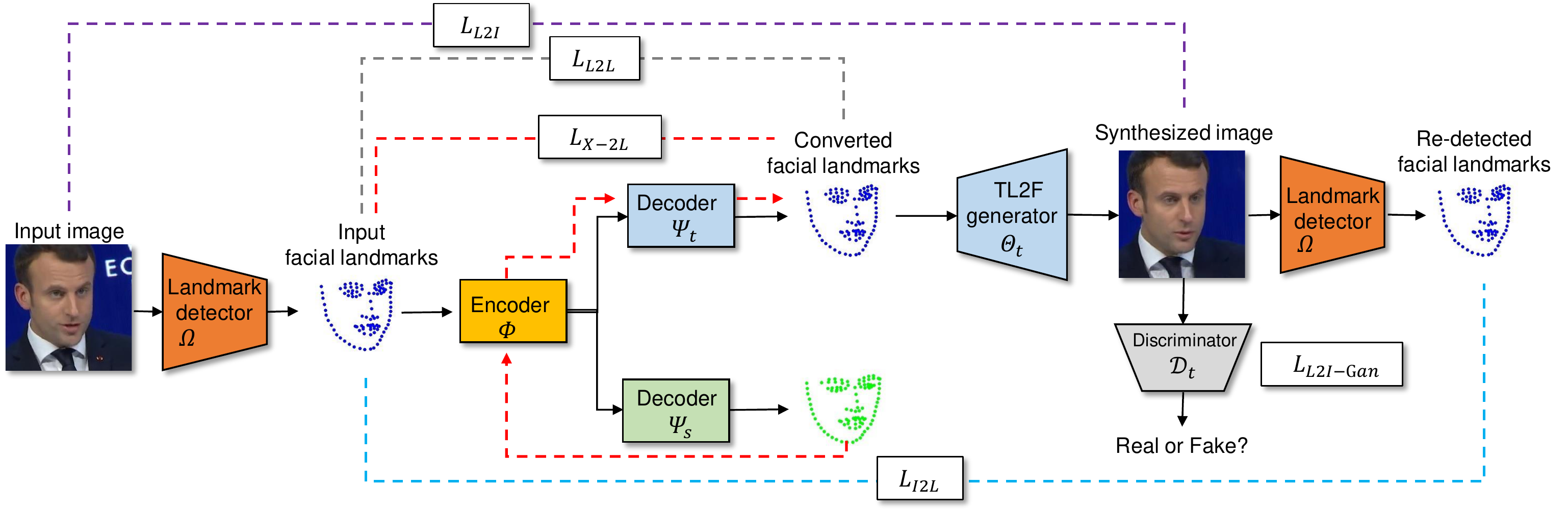}
    \vspace{-0.5cm}
    \caption{ \small Overview of training process of our method. See text for details.}
    \label{fig:training}
~\vspace{-1.7em} \end{figure*}

%%%%%%%%%%%%%%%%%%%%%%%%%%%%%%%%%%%%%%
\vspace{-0.3cm}
\subsection{Training}
\label{sec:training}
\vspace{-0.3cm}

As we do not assume correspondence in facial expressions among landmarks of different identity, the loss function is formed in a self-regularized manner.  The overall loss function is the sum of five loss terms, as
\begin{equation}
    L_\text{overall} = L_\text{L2I} +
    L_\text{I2L} +
    L_\text{L2L} + L_\text{X-L2L} + L_\text{L2I-gan}.
    \label{eq:overall}
\end{equation}

The first term corresponds to the $\ell_1$ error between an input image and its reconstruction from its landmarks using the target specific landmark-to-face generator. Specifically, for a target face $t$ with an input image $\I_t$ and the corresponding facial landmarks $\L_t$, the image reconstruction loss is given by  % $\L_t = \Omega(\I_t)$,  
\begin{equation}
    L_{\text{L2I}} = \mathbb{E}_{t} [\|  {\I}_{t} -  \Theta_{t}({\L}_{t})\|_1],
\end{equation}
where $\mathbb{E}_{t}$ denotes the average over all training identities. 

The second term ensures the facial landmarks are well preserved in synthesized face. Denote $\Omega$ as the landmark detector. This term can be written as 
\begin{equation}
    L_\text{I2L} = \mathbb{E}_{t} [\; ||{\L}_{t} -  \Omega(\Theta_{t}({\L}_{t}))||_2 \;].
    \label{eq:i2l}
\end{equation}

\noindent{\bf Differentiable landmark detector}. To be able to optimize this loss, a differentiable landmark detector is required. However, this is not the case for most existing state-of-the-art landmark detectors \cite{bulat2017far,sun2019deep,qian2019aggregation}, due to the non-differentiable argmax function used for final landmark selection. Therefore, we adapt the {\em differentiable spatial to numerical transform} (DSNT) module \cite{nibali2018numerical} to make a differentiable landmark detector. The DSNT converts the non-differentiable argmax function into a soft-argmax \cite{luvizon2017human} based function which can be differentiable.

The third term corresponds to the reconstruction of landmarks in the landmark converter, as 
\begin{equation}
L_{\text{L2L}} = \mathbb{E}_{t}[\;||{\L}_{t} -\Psi_t(\Phi({\L}_{t}))||_2 + || {\L^{'}_{t}} -  \Phi({\L}_{t}) ||_2\;],
\label{eq:l2l}
\end{equation}
where the first term of Eq.\eqref{eq:l2l} measures the landmark error between the input facial landmarks ${\L}_{t}$ and its reconstruction from the landmark converter, combining the general encoder $\Phi$ and target $t$'s decoder $\Psi_t$, $\L'_{t}$ in second term is the $\ell_2$ normalized landmarks $\L_{t}$, which guides the encoder to learn identity-independent landmarks in auxiliary.

The four term evaluates errors when landmarks of different identities' faces undergo with the landmark converter. Specifically, it is the difference between the landmark of identity $s$ and its cyclic transformation to and back from identity $t$, as
\begin{equation}
    L_\text{X-L2L} = \mathbb{E}_{t \neq s}[\;||{\L}_{t} - \Psi_{t}(\Phi(\Psi_{s}(\Phi({\L}_{t}))))||_2\;].
\end{equation}

The last term couples the training of the landmark converter and the target specific landmark to image generator, which is a landmark-to-image GAN loss as
\begin{equation}
\begin{array}{ll}
     L_{\text{L2I-gan}} = & \mathbb{E}_{t} [[\; \log {\CD}_{t} ({\I}_{t}) \;] + \\
     & \mathbb{E}_{s} [\; \log(1 - {\CD}_{t}(\Theta_{t}(\Psi_{t}(\Phi({\L}_{s}))) \;]],
\end{array}
    % L_{\text{L2I-gan}} = \mathbb{E}_{t} [[\; \log {\CD}_{t} ({\I}_{t}) \;] + \mathbb{E}_{s} [\; \log(1 - {\CD}_{t}(\Theta_{t}(\Psi_{t}(\Phi({\L}_{s}))) \;]].
\end{equation}
where ${\CD}_t$ denotes a discriminator as in the PatchGAN model \cite{liu2019few} to distinguish the original and synthesized face images of identity $t$. 

% As in the general GAN loss \cite{goodfellow2014generative}, $L_{\text{L2I-gan}}$ is the cross-entropy loss of the discriminator on original images of target $t$'s faces and those reconstructed from landmarks of arbitrary identities using the landmark converter and target-specific landmark-to-face generator.

% Training of the overall model proceeds as alternative updates of the parameters in the landmark converter and the target-specific landmark-to-face generator to minimize Eq.\eqref{eq:overall}. Specifically, we first extract landmarks from all training faces. Then, starting with initial parameter settings, we update the parameters in landmark converter, $\Phi$ and $\Psi_t$, fixing the parameters in the target-specific landmark-to-face generator $\Theta_t$. Then, we minimize the loss function with regards to parameters in $\Theta_t$ while fixing the parameters in $\Phi$ and $\Psi_t$. The two steps are repeated until converge or the maximum number of iteration is reached.

%%%%%%%%%%%%%%%%%%%%%%%%%%%%%%
\vspace{-0.3cm}
\section{Experiments}
\vspace{-0.3cm}

\subsection{Experimental Settings}
\vspace{-0.3cm}
\noindent{\bf Dataset.} We train and test our method using the CelebV dataset \cite{wu2018reenactgan}. We choose this dataset because it has been used in previous works \cite{wu2018reenactgan,ha2020marionette}. The CelebV dataset includes faces of five identities, namely, Emmanuel Macron, Kathleen, Jack Ma, Theresa May, and Donald Trump. 

\smallskip
\noindent{\bf Implementation details.} 
Our method is implemented using PyTorch 1.0.1 on Ubuntu 16.04 with a Nvidia 1080ti GPU.
The models are trained using the RMSProp optimizer \cite{RMSProp} with kaiming initialization \cite{he2015delving} of the model parameters. For the landmark converter, the training batch size is $4$, the learning rate starts as $10^{-5}$, and the maximum iteration is set to $45,000$. For the target-specific landmark-to-image generator, the batch size is set to $1$ and the maximum iteration is set to $4 \times 10^5$. The learning rate starts as $6 \times 10^{-5}$, and is decayed $10\%$ every $2,500$ iterations. 
For the differentiable landmark detector, we use one stacked hourglass structure as the base network to save the resource cost in training. Our landmark detector is trained on the WFLW dataset \cite{wayne2018lab}. 

% Under the evaluation metric NME\footnote{Normalized Mean Error (NME) is the average distance between detected landmarks and ground truth which is then normalized by the distance between the leftmost key point in left eye and the rightmost key point in right eye.}, the differentiable landmark detector can achieve score $0.07$, which is competitive to state-of-the-art methods \cite{sun2019deep,qian2019aggregation}, which are $0.05$ and $0.04$, respectively.

%%%%%%%%%%%%%%%%%%%%%%%%%%%%%%%%%%%
\vspace{-0.3cm}
\subsection{Landmark to Face Synthesis}
\label{sec:face_synthesis}
\vspace{-0.3cm}

%Our method achieve $0.63$ in SSIM and $0.59$ in LDIF.

% \begin{figure}[t]
%   \centering
%   \includegraphics[width=\linewidth]{figure/demo.pdf}
%   \vspace{-0.5cm}
%   \caption{ \small Examples of landmark to face synthesis using input and target identities from the CelebV dataset. }
%   \label{fig:demo1}
% ~\vspace{-2em} \end{figure}

% \begin{figure}[t]
%   \centering
%   \includegraphics[width=\linewidth]{figure/teaser.pdf}
%   \vspace{-0.7cm}
%   \caption{\small Examples of landmark to face synthesis. }
%   \label{fig:demo1}
% ~\vspace{-2em} \end{figure}

% Several visual examples of the proposed LandmarkGAN are shown in Fig.\ref{fig:demo1}. These results show that the facial expressions and orientations embodied by the landmarks are well reproduced in the synthesized images. 

\begin{figure}[t]
    \centering
    \includegraphics[width=\linewidth]{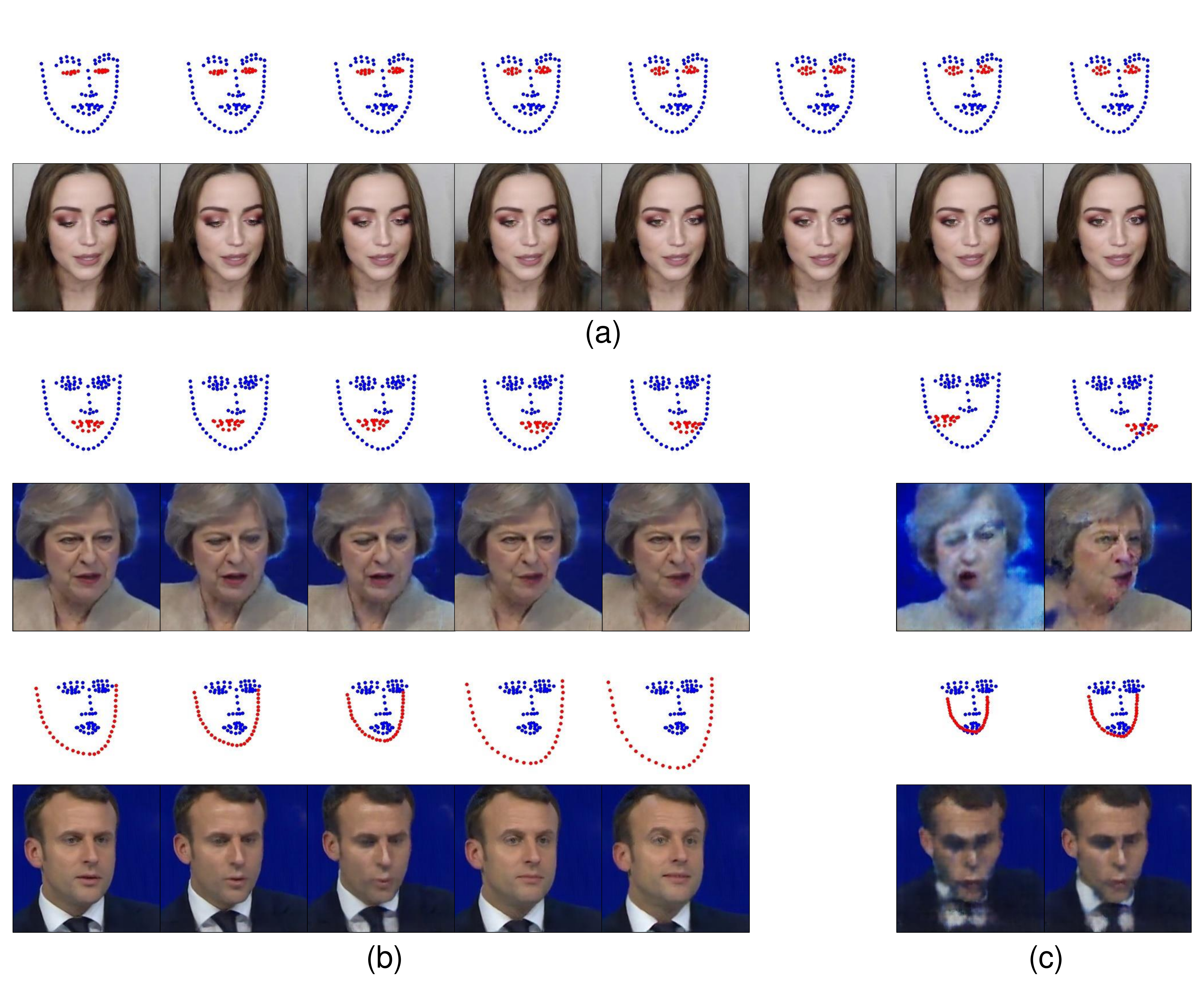}
    \vspace{-1cm}
    \caption{ \small Visual examples of face synthesis by editing landmarks.}
    \label{fig:face_edit}
~\vspace{-2em} \end{figure}

Fig. \ref{fig:teaser} shows several examples of landmark to face synthesis. To further demonstrate the flexibility of our method, we conduct another set of experiments to progressively change the landmarks corresponding to eyes, see Fig.\ref{fig:face_edit} (a). 
We further generate faces with more extreme editing of landmarks to see the response of our method. The first setting is we shift the location of mouth to an extreme location that is not existed in the training data in CelebV. In the second setting, we edit the face contour to change the shapes of the targets' faces. Fig.\ref{fig:face_edit} (b) shows the visual examples of our method to different inputs, where the top and bottom part correspond to two settings respectively.  As in the previous cases, we can observe that the edited facial landmarks lead to corresponding changes in the synthesized faces. Moreover, our method can adjust the synthesized face properly to keep the semantic meaning even though the shapes of the faces are changed. These confirm that our method are flexible to larger changes to the landmarks and can synthesize faces with variations that are not present in the training data.
However, our method still has the limit which can not handle the facial landmarks with extreme editing such as moving the mouth outside of face or largely shrinking the face contour. Fig.\ref{fig:face_edit} (c) shows several failure examples of our method. 

% [Add failure cases] 

%%%%%%%%%%%%%%%%%%%%%%%%%%%%%%%%%%%%%%%
\vspace{-0.5cm}
\subsection{Face Reenactment}
\vspace{-0.3cm}

In the second set of experiments, we build a face reenactment system based on our landmark to face synthesis method by adding a facial landmark extractor. Specifically, given an input face image, we extract facial landmarks, which are then fed to create a face preserving the facial expressions and orientations. 

\begin{figure}[t]
  \centering
%   \vspace{-0.2cm}
  \includegraphics[width=\linewidth]{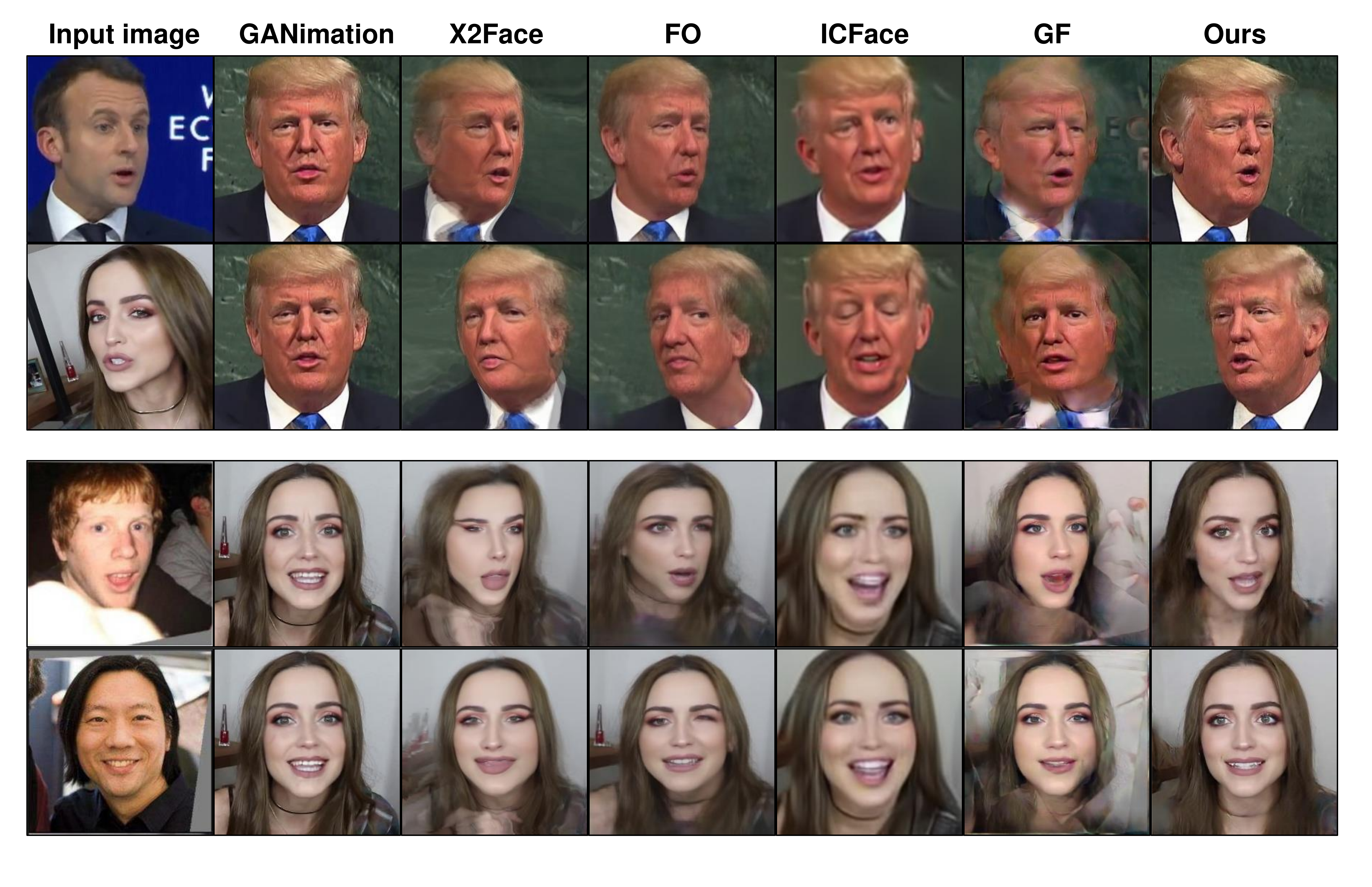}
   \vspace{-1cm}
  \caption{ \small Qualitative comparison of each method on CelebV dataset (left) and wild images (right). See text for details.}
  \label{fig:vis_comp}
~\vspace{-2em} \end{figure}

% \begin{figure}[t]
%   \centering
% %   \vspace{-0.2cm}
%   \includegraphics[width=1.\linewidth]{figure/vis_comp_2.pdf}
%   \vspace{-0.8cm}
%   \caption{ \small Qualitative comparison of each method on wild images. }
%   \label{fig:vis_comp_2}
% ~\vspace{-2em} \end{figure}

\begin{table}[t]
	\centering
	\footnotesize
	\caption{\small Quantitative evaluations. See text for details.}
% 	\vspace{-0.3cm}

	\begin{tabular}{|c|c|c|c|}
		\hline	
		Methods  & LMK$\downarrow$ & SSIM$\uparrow$ & ID$\downarrow$ \\
% % 		\hline
% % 		\hline 
% %         CycleGAN$^*$ & 1.67 & 0.60 & \bf 0.37 \\
% % 		\hline
% % 		\bf LandmarkGAN$^*$ & \bf 0.67 & \bf 0.67 & 0.39 \\
% % 		\hline
% % 		\hline
% % 		ReenactGAN$^*$ & 1.15 & 0.54 & \bf 0.35 \\
% % 		\hline
% % 		\bf LandmarkGAN$^*$ & \bf 0.57 & \bf 0.63 & 0.39 \\
% % 		\hline
% 		\multicolumn{4}{c}{} \\
		\hline
		GANimation & 3.09 & 0.46 & 0.37 \\
		\hline
		X2Face & 1.14 & 0.61 & 0.45 \\
		\hline
		FO & 1.02 & 0.63 & 0.40 \\
		\hline
		ICFace & 3.33 & 0.47 & 0.47 \\
		\hline
		GF & 1.63 & 0.55 & 0.41 \\
% 		\hline
%         \lyz{FSGAN} & \ly{11.77} & \lyz{0.44} & \lyz{0.61} \\
		\hline
		\hline
		\bf LandmarkGAN & \bf 0.77 & \bf 0.68 & \bf 0.31 \\
		\hline
	\end{tabular}
	\label{table:eval}
	\vspace{-0.7cm}
\end{table}

% \begin{figure}[t]
%   \centering
%   \includegraphics[width=\linewidth]{figure/wildface_demo.pdf}
%   \vspace{-0.8cm}
%   \caption{ \small Examples of our method on input faces not part of the CelebV dataset (unseen images).}
%   \label{fig:wildface_comp}
% ~\vspace{-2em} \end{figure}

\smallskip
\noindent{\bf Compared methods.} 
We compare with five state-of-the-art methods using input and target identities from the CelebV dataset, which are {\bf GANimation} \cite{pumarola2018ganimation}, {\bf X2Face} \cite{wiles2018x2face}, {\bf FO} \cite{siarohin2019first}, {\bf ICFace} \cite{tripathy2020icface} and {\bf GF} \cite{ren2020deep}. Specifically, GANimation achieves facial expression synthesis based on Action Units (AU) annotations from a single image\footnote{Note the input of GANimation is the central face instead of the whole face. For comparison with others, we paste the synthesized face area back to the same location in original image.}. X2Face is a self-supervised network that can transfer the pose and expression of a source face to a target face. Similar to GANimation, FO also achieves the face animation by decoupling the appearance and motion information using a self-supervised formulation. ICFace achieves face reenactment using human interpretable control signals such as head pose angles and AU values. GF is designed for pose-guided person image generation using global-flow local-attention model. 

% For comparison, we retrain CycleGAN on the CelebV dataset. Since training CycleGAN on each pair of identities is time-consuming, we only train models using Trump as the target identity. For ReeanctGAN, we follow the instructions to train it on the CelebV dataset. However, the quality of results are not competitive as in original paper. Therefore we directly use the pretrained model of synthesizing Trump provided by the authors. Since the other methods are disentanglement based, we directly use the published models for comparison.}

\smallskip
\noindent{\bf Visual comparison.} 
{Fig.\ref{fig:vis_comp} shows qualitative comparison of each method on CelebV dataset (top) and wild images (bottom), which shows our method is better at preserving facial expressions and head orientations. Note for disentanglement based methods, we follow the instructions of each method to select the reference images of target identity and utilize the input images as driven images.}

\smallskip
\noindent{\bf Quantitative evaluations.} 
We next compare results of face reenactment quantitatively using three metrics: {\em landmark difference (LMK)}, {\em SSIM}, {\em identity difference between synthesized and target face (ID)}. 
LMK aims to evaluate the fidelity of the synthesized face images in terms of preserving the landmarks of the input image. Since the identity is changed during face reenactment, directly evaluating the landmarks preserving between the source and target identity is difficult due to the large variety of face shape in different identity. To this end, we use the synthesized target face of each method as input again to synthesize the source face. In this way, we first normalize the landmarks to $[0, 1]$ and then calculate the $\ell_2$ landmark difference between the original source face and synthesized source face. 
% The average LMK score is shown in Table \ref{table:eval}. As these results show, reenacted faces using our method have aligned facial landmarks more similar to those in the input faces, which suggests that facial expressions and orientations are better preserved. 
SSIM \cite{wang2004image} is used in recent works  \cite{ha2020marionette,zhang2020freenet,zakharov2019few} to evaluate visual quality. Since the ground truth of synthesized target face is not existed in our experiment and it is not appropriate to directly calculate the SSIM score of synthesized target face referred on the source face, we use the similar setting as in LMK. 
% Specifically, the SSIM score of CycleGAN and our method to synthesize Trump's face using other four input identities are $0.60$, $0.67$ respectively. To compare ReenactGAN with our method, we compute the SSIM by using the synthesized image from the generator which has the same identity with input image, where our method is $0.63$ and ReenactGAN is $0.54$. The SSIM score of other methods are shown in Table \ref{table:eval}. The results reveal that our method achieves better visual quality compared to others. Visual examples of using this metric is shown in supplementary material.
We then evaluate the quality of identity swap. Specifically, we select a frontal face of the target identity from CelebV dataset as reference for each method, then we calculate the face recognition score (\eg, Dlib \cite{dlib09}) between the reference and synthesized face of target. Table \ref{table:eval} shows the details of quantitative evaluation, which demonstrates the effectiveness of our method.

\vspace{-0.3cm}
\subsection{Ablation Study of Landmark Converter}
\begin{figure}[t]
  \centering
  \includegraphics[width=0.9\linewidth]{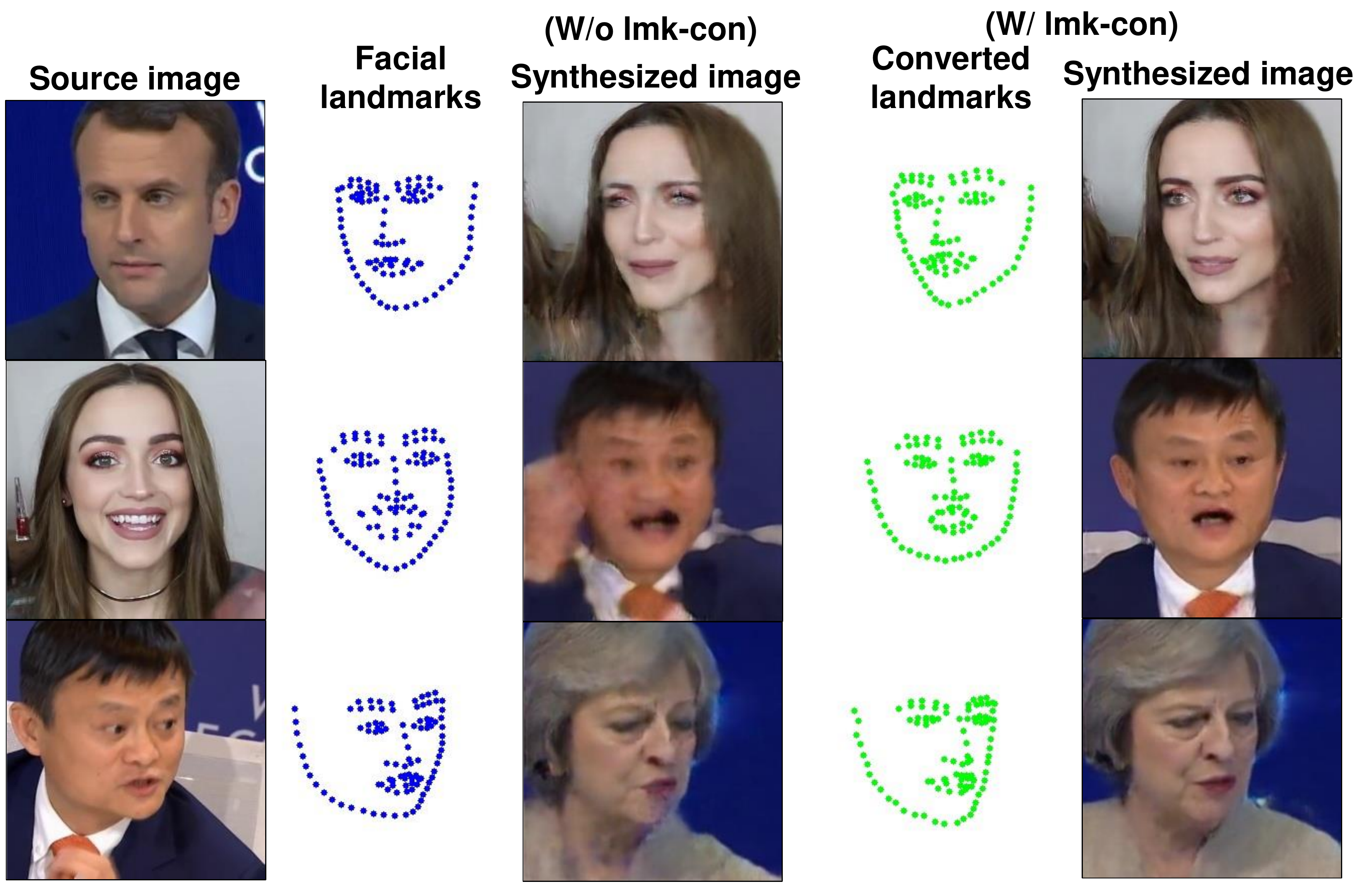}
  \vspace{-0.5cm}
  \caption{ \small Ablation study on the effectiveness of landmark converter.}
  \label{fig:ablation_wo_LT}
~\vspace{-2em} \end{figure}

\vspace{-0.3cm}
%We conduct further experiments to demonstrate the effectiveness of the landmark converter and the influence of individual terms in loss function in the overall system. 
% \smallskip
% \noindent{\bf Effectiveness of landmark converter.} 
To demonstrate the role of landmark converter, we conduct a set of experiments by directly feeding the extracted facial landmarks without using landmark converter. Fig.\ref{fig:ablation_wo_LT} shows the results without ({\tt w/o lmk-con}) and with landmark converter ({\tt w/ lmk-con}). The quality of synthesized faces is significantly degraded without using the landmark converter, exhibited as blurring and inaccurate face shapes. The fourth and fifth column are the landmarks after landmark converter and corresponding synthesized results. It reveals the output of landmark converter has an intuitive deformation, which adjusts landmarks to better fit the target identity.

% \begin{figure}[h]
%   \centering
%   \includegraphics[width=0.24\linewidth]{figure/T1_abl.pdf}
%   \includegraphics[width=0.24\linewidth]{figure/T2_abl.pdf} 
%   \includegraphics[width=0.24\linewidth]{figure/T3_abl.pdf} 
%   \includegraphics[width=0.24\linewidth]{figure/T4_abl.pdf} \\
%   \includegraphics[width=0.24\linewidth]{figure/T1_abl_bar.pdf}
%   \includegraphics[width=0.24\linewidth]{figure/T2_abl_bar.pdf} 
%   \includegraphics[width=0.24\linewidth]{figure/T3_abl_bar.pdf} 
%   \includegraphics[width=0.24\linewidth]{figure/T4_abl_bar.pdf}
%   \vspace{-0.4cm}
%   \caption{ \small Ablation study on the performance of AUDIF and LDIF on w/o $L_\text{I2L}$, w/o $L_\text{L2I-gan}$ and original setting.}
%   \label{fig:ablation}
% ~\vspace{-2em} \end{figure}

%%%%%%%%%%%%%%%%%%%%%%%%%%%%%%
%%%%%%%%%%%%%%%%%%%%%%%%%%%%%%
\vspace{-0.3cm}
\section{Conclusions}
\vspace{-0.3cm}

In this work, we describe a new method, known as LandmarkGAN, to synthesize faces from facial landmarks. Facial landmarks are a natural, intuitive, and effective representation for facial expressions and orientations, which are independent from the target's texture or color and background scene. Our model consists of two components: a landmark converter which converts the input facial landmarks to those of target face, and a target-specific landmark-to-face generator which synthesize a target face based on converted facial landmarks. Face synthesis and reenactment experiments conducted on CelebV dataset demonstrate the effectiveness of our method. 

% For future works, we would like to study face synthesis and reenactment method based on 3D facial landmarks. Moreover, we will also consider temporal consistency explicitly in our method to enhance the naturalness in face synthesis and reenactment videos.

% For future works, we would like to extend our model in the following directions. First, we would like to reduce the number of training face images and extend the current method to the one-shot or few-shots settings. In addition, 3D facial landmarks can provide more information about facial expressions and orientations, and in the next step we will study face synthesis and reenactment method based on 3D facial landmarks. We will also consider temporal consistency explicitly in our method to create natural-looking face synthesis and reenactment videos.

\small{
\bibliographystyle{IEEEbib}
\bibliography{egbib}
}
\end{document}